\pdfoutput=1

\documentclass[11pt]{article}

\usepackage[final]{acl}

\usepackage{times}
\usepackage{latexsym}

\usepackage[T1]{fontenc}

\usepackage[utf8]{inputenc}

\usepackage{microtype}

\usepackage{inconsolata}
\usepackage{graphicx}
\usepackage{tabularx}
\usepackage{soul}

\usepackage{multirow}
\usepackage{amsmath,amssymb,amsfonts}
\usepackage{amsthm}
\usepackage{mathrsfs}
\usepackage[title]{appendix}%
\usepackage{xcolor}%
\usepackage{textcomp}%
\usepackage{manyfoot}%
\usepackage{booktabs}%
\usepackage{algorithm}%
\usepackage{algorithmicx}%
\usepackage{algpseudocode}%
\usepackage{listings}%
\usepackage{float}
\usepackage{makecell}
\usepackage{ulem}

\title{EvidenceMap: Learning Evidence Analysis to Unleash the Power of Small Language Models for Biomedical Question Answering}

\makeatletter
\def\@fnsymbol#1{\ensuremath{\ifcase#1\or \or \ddagger\or
\mathsection\or \mathparagraph\or \|\or **\or \dagger\dagger
\or \ddagger\ddagger \else\@ctrerr\fi}}
\makeatother

\author{Chang Zong$^1$, Jian Wan$^1$, Siliang Tang$^2$, Lei Zhang$^1$${^\dagger}$ \\ $^1$Zhejiang University of Science and Technology \\
$^2$Zhejiang University \\
\texttt{zongchang@zust.edu.cn}
}

\begin{document}
\maketitle
\begin{abstract}
When addressing professional questions in the biomedical domain, humans typically acquire multiple pieces of information as evidence and engage in multifaceted analysis to provide high-quality answers. Current LLM-based question answering methods lack a detailed definition and learning process for evidence analysis, leading to the risk of error propagation and hallucinations while using evidence. Although increasing the parameter size of LLMs can alleviate these issues, it also presents challenges in training and deployment with limited resources. In this study, we propose \textbf{\texttt{EvidenceMap}}, which aims to enable a tiny pre-trained language model to explicitly learn multiple aspects of biomedical evidence, including supportive evaluation, logical correlation and content summarization, thereby latently guiding a small generative model (around 3B parameters) to provide textual responses. Experimental results demonstrate that our method, learning evidence analysis by fine-tuning a model with only 66M parameters, exceeds the RAG method with an 8B LLM by 19.9\% and 5.7\% in reference-based quality and accuracy, respectively.
\end{abstract}

\section{Introduction}

\begin{figure}[htbp]
    \begin{center}
    \includegraphics[width=0.48\textwidth]{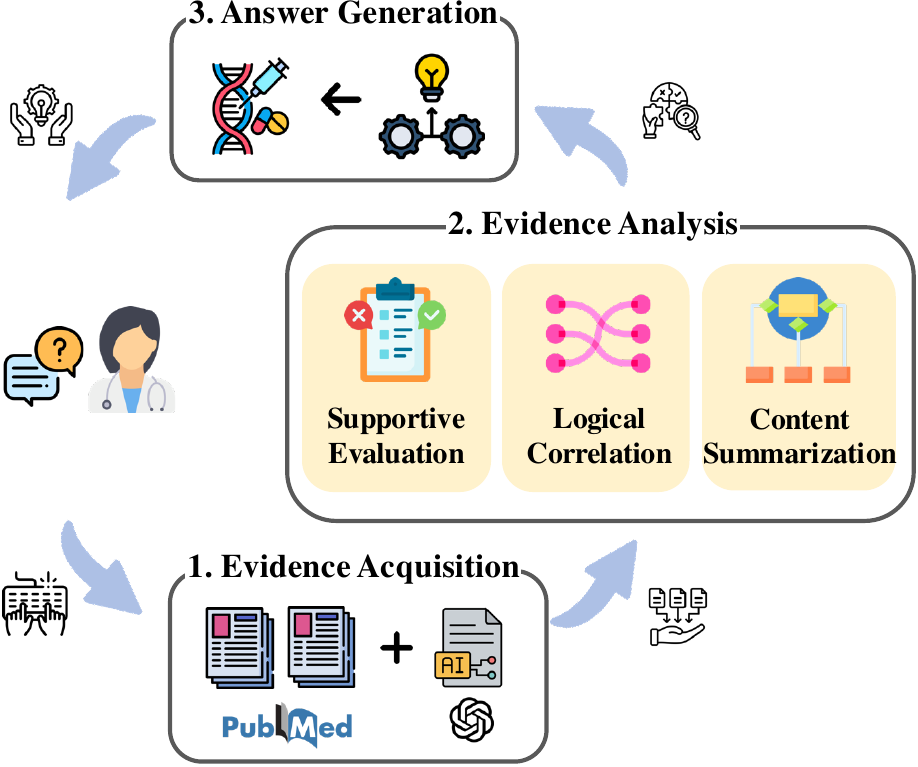} 
    \caption{A diagram describing the analysis with various evidence when addressing biomedical questions.}
    \label{diagram}
    \end{center}
\end{figure}

Answering professional questions typically requires gathering a wide range of information as evidence, such as relevant academic articles and domain knowledge from experts \cite{louis2024interpretable,singhal2025toward}. In the biomedical domain, it is crucial to formulate answers by evaluating and summarizing various evidence \cite{singhal2023large,zhang2024closing}. We describe a modern scenario in Figure \ref{diagram}, in which a biomedical expert acquires relevant evidence from PubMed papers and a large language model (LLM) and performs a series of analytical steps that include supportive evaluation \cite{jin2021disease}, logical correlation \cite{chen2021logical}, and content summarization \cite{xie2022pre}, ultimately delivering a high quality answer.

\begin{figure}[htbp]
    \begin{center}
    \includegraphics[width=0.48\textwidth]{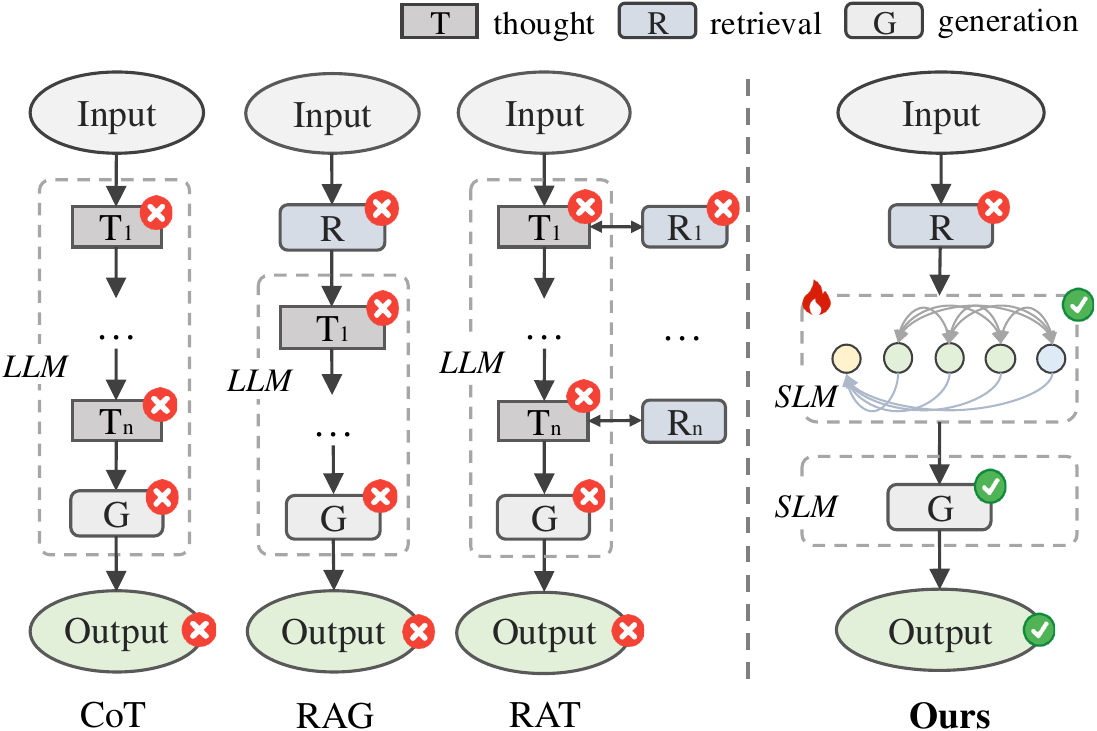} 
    \caption{Existing methods (left) suffer from erroneous thinking process or reasoning on low-quality evidence. Our framework (right) reduces error propagation by learning evidence analysis explicitly and efficiently with small language models (SLM).}
    \label{paradigm}
    \end{center}
\end{figure}

To answer domain questions, LLM-based generation methods have shown advantages in recent years. Current methods primarily incorporate external knowledge into LLMs for reasoning \cite{lewis2020retrieval,asai2023self,he2024g,wang2024rat} or enable LLMs to generate thinking processes \cite{wei2022chain,lightman2023let,yao2024tree}, thus mitigating hallucinations and producing accurate and comprehensive responses. However, as illustrated in Figure \ref{paradigm}, due to a lack of training in analysis skills and descriptions of evidence interactions, these approaches still face the issue of error propagation when addressing biomedical questions. Moreover, although increasing the scale of model parameters can improve reasoning capabilities, it poses challenges for training and deployment under limited resources.

To alleviate the aforementioned issues, this study focuses on biomedical question answering with various evidence, and proposes a novel framework, named \textbf{\texttt{EvidenceMap}}. The framework explicitly learns evidence analysis with small language models, thereby efficiently mimicking the human-solving process in the biomedical domain. Specifically, our framework defines a conceptual model called Evidence Map, which describes the relational dynamics among multiple pieces of evidence relevant to the question, and subsequently employs a small pre-trained language model (PLM) to learn the representation of analytical elements in the Evidence Map, including supportive evaluation, logical correlation, and content summarization, which serve as the prompt to drive a small generative language model (GLM) to produce answers. The experimental results indicate that learning evidence analysis can significantly enhance the performance of biomedical question answering. Our contributions are summarized as follows.

\begin{itemize}
    \item We present \textbf{\texttt{EvidenceMap}}, a novel question answering framework inspired by biomedical evidence analysis, which implements multifaceted analysis including supportive evaluation, logical correlation, and content summarization in multiple pieces of evidence. 
    \item We develop an efficient evidence analysis learning method that utilizes semantic abilities of small pre-trained language models to obtain representations of various components of analysis, thereby guiding generative language models to produce the final answer.
    \item Extensive experiments and analyses on biomedical datasets demonstrate that our approach, by explicitly learning and utilizing evidence analysis, can significantly improve the quality of answers and outperforms reasoning methods with larger-scale language models.
\end{itemize}

\section{Definitions}

\subsection{Evidence Map Modeling} \label{section2.1}
We define an evidence map ($\mathcal{M}$) for each input question to facilitate reasoning during generation. An evidence map introduces three key types of information, including the evaluation of how the evidence supports the question ($\mathcal{R}_{eval}$), the correlation between the evidence ($\mathcal{R}_{cor}$), and the summarization of all the evidence ($E_{sum}$). Generally, an evidence map can be formulated as follows.

\begin{equation}
\begin{aligned}
\mathcal{M} &= \{Q, \mathcal{E}_{1}, \mathcal{E}_{2}, ... \mathcal{E}_{m}, \mathcal{R}_{eval}, \mathcal{R}_{cor}, E_{sum}\}, \\
E_{sum} &= Agg(\mathcal{E}_{1}, \mathcal{E}_{2}, ... \mathcal{E}_{m}),
\label{eq:eq1}
\end{aligned}
\end{equation}where $Q$ is the question to be solved, $Agg(\cdot)$ is the aggregation function to derive the summarization $E_{sum}$, $\mathcal{E}_{s}$ represents a set of evidence from the source $s$, which is denoted as follows.

\begin{equation}
\begin{aligned}
\mathcal{E}_{s} &= \{E_{s}^1, E_{s}^2, ..., E_{s}^n\},
\label{eq:eq2}
\end{aligned}
\end{equation}where $E_{s}^i$ is the $i$-th evidence from the source $s$. Furthermore, the supportive evaluation and logical correlation of evidence can be described as below.

\begin{equation}
\begin{aligned}
R_{eval}^{i} &= [E_s^i, Q, \rightarrow], \forall R_{eval}^{i} \in \mathcal{R}_{eval}, \\
R_{cor}^{(i,j)} &= [E_s^i, E_s^j, \leftrightarrow], \forall R_{cor}^{(i,j)} \in \mathcal{R}_{cor},
\label{eq:eq3}
\end{aligned}
\end{equation}where the symbol $\rightarrow$ and $\leftrightarrow$ represent the direction of these two relationships, respectively.

\subsection{Learning Generative QA with LMs}
The objective of our task is to generate the best answers to biomedical questions by learning evidence analysis from the evidence map, which can be described as follows.

\begin{equation}
\begin{aligned}
A^* = f_{glm} \big ( Q, f_{plm}(\mathcal{M};\theta^*) \big ),
\label{eq:eq4}
\end{aligned}
\end{equation}where $f_{plm}(\cdot)$ is a tiny pre-trained model (e.g. DistilBERT-Base), $f_{glm}(\cdot)$ is a small generative model (e.g. Llama-3.2-3B) for providing the best answer $A^*$. Then, our learning process is to find the optimized parameter set $\theta^*$, which leads to a minimum generative loss for all question-answer pairs, given the corresponding evidence map $\mathcal{M}_i$ for each question $Q_i$. This can be denoted as below.

\begin{equation}
\begin{aligned}
\theta^* = \mathop{\arg\min}\limits_\theta \sum_i \mathcal{L} \Big(A_i, f_{glm} \big(Q_i, f_{plm}(\mathcal{M}_i;\theta) \big) \Big).
\label{eq:eq5}
\end{aligned}
\end{equation}

\begin{figure*}[htbp]
    \begin{center}
    \includegraphics[width=1\textwidth]{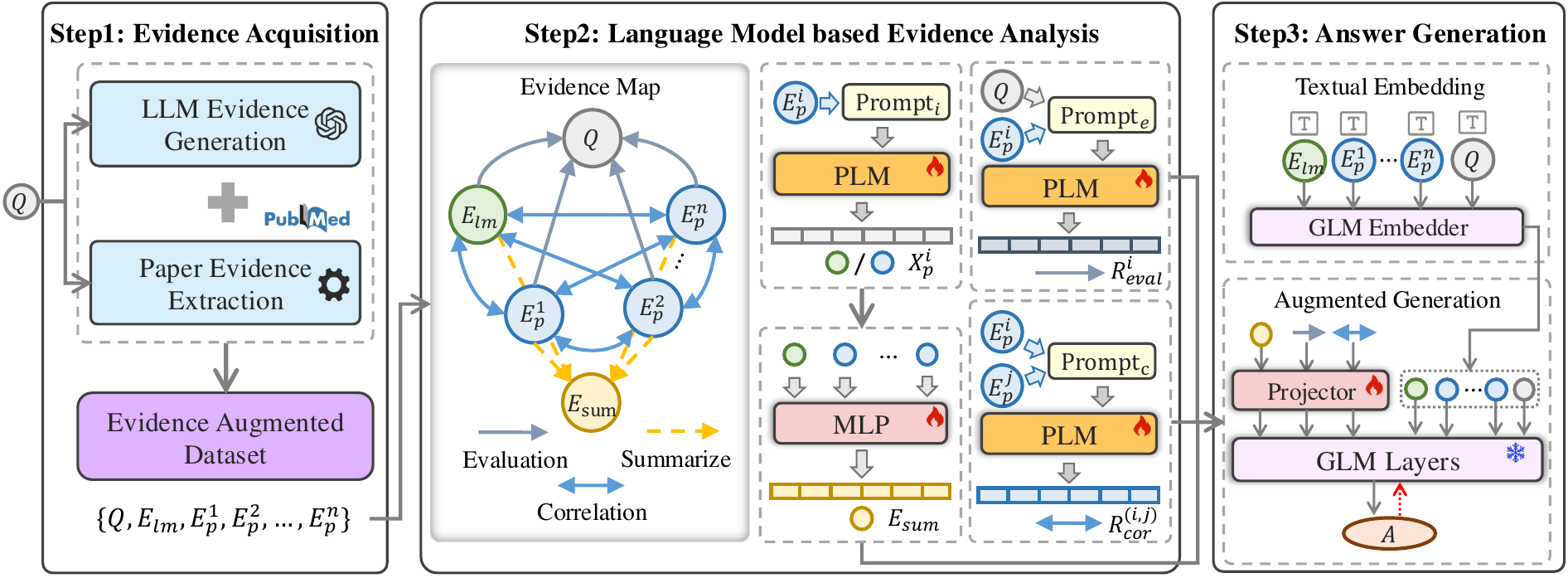} 
    \caption{The overview of our \textbf{\texttt{EvidenceMap}} framework. An evidence-augmented question is modeled as an evidence map, in which the representations of evidence evaluation, correlation and summarization are computed by a pre-trained language model (PLM) accordingly, in order to facilitate answering by a generative model (GLM).}
    \label{framework}
    \end{center}
\end{figure*}

\section{Framework}
Our \textbf{\texttt{EvidenceMap}} framework is illustrated in Figure \ref{framework}, in which a pre-trained language model is fine-tuned to learn multifaceted evidence analysis to facilitate responses with a small generative model. The three main steps are described below.

\subsection{Evidence Acquisition} \label{acquisition}
As described in Figure \ref{diagram}, we gather relevant evidence from multiple sources, including a powerful commercial LLM and academic papers, for each question. Specifically, for evidence from an LLM, we ask GPT-4o \cite{hurst2024gpt}, a powerful commercial LLM, to utilize its own knowledge to provide a series of key points that support the question, which are described concisely (see Appendix \ref{prompt-acquire}). In addition, we directly extract paper snippets from existing biomedical QA datasets such as BioASQ \cite{krithara2023bioasq} and PubMedQA \cite{pubmedqa}, which have already been annotated by human experts, as the evidence related to the question. Therefore, an evidence-augmented QA dataset is constructed, in which each question corresponds to one piece of LLM evidence $E_{lm}$ and multiple pieces of paper evidence from $E_p^1$ to $E_p^n$. The input data can be described as follows.

\begin{equation}
\begin{aligned}
Input = \{Q, E_{lm}, E_p^1, E_p^2, \cdots , E_p^n\}.
\label{eq:eq6}
\end{aligned}
\end{equation}

\subsection{LM-based Evidence Analysis}
\paragraph{Evidence Map.} As defined in Section \ref{section2.1}, each input question, along with its associated evidence, can be modeled as an evidence map to facilitate reasoning. In addition to the evidence itself, an evidence map also incorporates multifaceted information on analysis, including the evaluation of how each evidence supports the question $R_{eval}^i$, the correlation between any two pieces of evidence $R_{cor}^{(i,j)}$, and the summarization of all the evidence $E_{sum}$.

\paragraph{Evidence Analysis.} We prompt a small pre-trained language model to perform the multifaceted evidence analysis in latent space. Specifically, for evidence summarization, we first obtain the overall semantics of each evidence node, and then use a multilayer perceptron (MLP) to aggregate and produce a summarized representation of the evidence, which is formulated as below.

\begin{equation}
\begin{aligned}
H_s^{(1)} &= W_s^{(1)} \cdot X + b_s^{(1)}, \\ 
E_{sum} &= ReLU(H_s^{(n)}),
\label{eq:eq7}
\end{aligned}
\end{equation}where $X$ is the concatenated feature of all evidence nodes, denoted as $[X_{lm}, X_p^1, \cdots , X_p^n]$, $H_s^{(n)}$ is the embedding of the $n$-th layer. Inspired by PromptBERT \cite{promptbert-2022} and PromptEOL  \cite{scaling-sentence}, we prompt the pre-trained language model $f_{plm}$ and get the last hidden states of the model as the evidence feature $X_p^i$, which is described as follows.

\begin{equation}
\begin{aligned}
X_p^i = f_{plm}^{-1}(E_p^i, \texttt{Prompt}_i; \theta).
\label{eq:eq8}
\end{aligned}
\end{equation}Meanwhile, the evaluation of how an evidence node supports the question $Q$ is derived from the same language model by using $\texttt{Prompt}_e$ as below.

\begin{equation}
\begin{aligned}
R_{eval}^i = f_{plm}^{-1}(E_p^i, Q, \texttt{Prompt}_e; \theta),
\label{eq:eq9}
\end{aligned}
\end{equation}where $R_{eval}^i$ indicates the supportive relationship between the evidence $E_p^i$ and the question $Q$. Similarly, the correlation between any two evidence nodes $E_p^i$ and $E_p^j$ can be formulated as below.

\begin{equation}
\begin{aligned}
R_{cor}^{(i,j)} = f_{plm}^{-1}(E_p^i, E_p^j, \texttt{Prompt}_c; \theta),
\label{eq:eq10}
\end{aligned}
\end{equation}which indicates the implicit logical relationships among evidence nodes. More details are provided in Appendix \ref{prompt-evidence}.

\subsection{Analysis-Augmented Generation}
With the result of the analysis from an evidence map, the final answer to the question is provided in an autoregressive way using a generative language model $f_{glm}$. We first convert all textual evidence and the question into an embedding $H_t$ using the first embedding layer of $f_{glm}$ as follows.

\begin{equation}
\begin{aligned}
H_t &= \text{Embed}([E_{lm}, E_p^1, \cdots, E_p^n, Q]).
\label{eq:eq11}
\end{aligned}
\end{equation} Then the analysis results are projected into the same space as the textual embedding to serve as a soft prompt, using a MLP network with each linear layer formualted as below.

\begin{equation}
\begin{aligned}
\hat{H}_a = \hat{W} \cdot  [E_{sum}, \mathcal{R}_{eval}, \mathcal{R}_{cor}] + \hat{b},
\label{eq:eq12}
\end{aligned}
\end{equation} where $R_{eval}^i \in \mathcal{R}_{eval}$ and $R_{cor}^{(i,j)} \in \mathcal{R}_{cor}$ are the sets of all the supportive and logical relationships, respectively. We then prepend the analysis embedding $\hat{H}_a$ to the textual embedding $H_t$ as the overall prompt to implement an analysis-augmented generation with the frozen language model $f_{glm}$, which is formulated as below.

\begin{equation}
\begin{aligned}
A = f_{glm}(\texttt{Prompt}_{all}; \theta^*),
\label{eq:eq13}
\end{aligned}
\end{equation} where $\texttt{Prompt}_{all} = [\hat{H}_a, H_t]$, and $\theta^*$ is the frozen parameter set of $f_{glm}$.

\subsection{Training Objective}
The autoregressive training objective concentrates on updating the set of parameters $\theta$ of a small pre-trained language model to provide the representation of evidence analysis to predict subsequent tokens accurately using a generative model. We calculate the loss with the probability of generating the $i$-th token of the target answer $A$ with the overall prompt and all previous tokens as follows.

\begin{equation}
\begin{aligned}
\mathcal{L} = \sum_{i=1}^L log \big(A_i | \texttt{Prompt}_{all}, A_{0:i-1}; \theta \big),
\label{eq:eq14}
\end{aligned}
\end{equation} where $L$ is the length of the target answer $A$.

\section{Evaluation}

\subsection{Implementation Details} \label{details}
In the step of learning evidence analysis, we employ DistilBERT \cite{sanh2019distilbert}, a lighter but faster model distilled from BERT \cite{bert-2019}, as the pre-trained model to be updated. In the generation step, the 1B and 3B versions of Llama-3.2 \cite{dubey2024llama} are used to provide answers. We conduct experiments on two popular biomedical datasets, BioASQ \cite{krithara2023bioasq} and PubMedQA \cite{pubmedqa}. During training, we set the learning rate as 5e-4, the batch size as 4, and the maximum number of paper evidence per question as 5. For each experiment, we train the language model for 10 epochs in a single NVIDIA A6000 GPU and evaluate the model on test sets.

\subsection{Evaluation Metrics}
Reference-based quality aims to evaluate the similarity between generated responses and reference answers, encompassing aspects of linguistic structure and semantics. ROUGE-L \cite{rouge-l} and BERTScore (denoted as BERT-S) \cite{zhang2019bertscore} are popular metrics within this category. LLM-based accuracy (denoted as LLM-ACC) addresses the shortcomings of traditional metrics in assessing precision. Based on existing work such as RAGAS \cite{ragas-2024} and Self-Evaluation \cite{calibrating-2024}, we prompt GPT-4o and also provide reference answers to assess the generated content, balancing accuracy and overall quality. Additional details are provided in Appendix \ref{prompt-llm}.

\subsection{Main Results}
In order to demonstrate the effectiveness of learning evidence analysis, we consider three configurations: \textit{1) Inference-LLM}: Using larger language models in general and biomedical domains, including Llama-3.1 \cite{dubey2024llama}, GPT-4o-mini \cite{hurst2024gpt}, and OpenBioLLM \cite{pal2024openbiollms}. \textit{2) Inference-SLM}: Using a smaller frozen generative model with various reasoning strategies, such as RAG \cite{lewis2020retrieval,gao2023retrieval}, CoT \cite{wei2022chain} and RAT \cite{wang2024rat}; \textit{3) Tuning-SLM}: Adapting the task by fine-tuning a small generative model with LoRA \cite{hu2021lora} or updating a small pre-trained model via a soft prompt tuning style \cite{prompt-tuning,prefix-tuning}. More details on baseline settings are provided in Appendix \ref{baseline}.

\begin{table*}[htbp]
\renewcommand{\arraystretch}{1.2}
\footnotesize
\centering
\begin{tabular}{l|lccccccc}
\bottomrule
\multirow{2}{*}{\textbf{Setting}} & \multirow{2}{*}{\textbf{Method}}&\multirow{2}{*}{\textbf{Size}}&\multicolumn{3}{c}{\textbf{BioASQ}} & \multicolumn{3}{c}{\textbf{PubMedQA}}
\\ 
& \multicolumn{1}{l}{} & \multicolumn{1}{l}{} & \multicolumn{1}{c}{ROUGE-L} & \multicolumn{1}{c}{BERT-S} & \multicolumn{1}{c}{LLM-ACC} & \multicolumn{1}{c}{ROUGE-L} & \multicolumn{1}{c}{BERT-S} & \multicolumn{1}{c}{LLM-ACC}
\\ \hline   
{} & \multirow{1}{*}{4o-mini-RAG} & \multirow{1}{*}{-} & \multicolumn{1}{c}{0.1549} & \multicolumn{1}{c}{0.5848} & \multicolumn{1}{c}{0.7039} & \multicolumn{1}{c}{0.1328} & \multicolumn{1}{c}{0.5501} & \multicolumn{1}{c}{0.5874}
\\
{Inference-LLM} & \multirow{1}{*}{Llama3.1-RAG} & \multirow{1}{*}{8B} & \multicolumn{1}{c}{0.1936} & \multicolumn{1}{c}{0.6077} & \multicolumn{1}{c}{0.7101} & \multicolumn{1}{c}{0.1835} & \multicolumn{1}{c}{0.6431} & \multicolumn{1}{c}{\underline{0.6440}}
\\
{} & \multirow{1}{*}{OpenBioLLM} & \multirow{1}{*}{8B} & \multicolumn{1}{c}{0.2196} & \multicolumn{1}{c}{0.5981} & \multicolumn{1}{c}{\underline{0.7355}} & \multicolumn{1}{c}{0.1936} & \multicolumn{1}{c}{0.5726} & \multicolumn{1}{c}{0.6276}
\\ \hline 
{} & \multirow{1}{*}{Llama3.2-RAG} & \multirow{1}{*}{3B} & \multicolumn{1}{c}{0.2956} & \multicolumn{1}{c}{0.6225} & \multicolumn{1}{c}{0.6843} & \multicolumn{1}{c}{0.1942} & \multicolumn{1}{c}{0.6102} & \multicolumn{1}{c}{0.6361}
\\
{Inference-SLM} & \multirow{1}{*}{Analysis-CoT} & \multirow{1}{*}{3B} & \multicolumn{1}{c}{0.2519} & \multicolumn{1}{c}{0.6044} & \multicolumn{1}{c}{0.6231} & \multicolumn{1}{c}{0.1918} & \multicolumn{1}{c}{0.6053} & \multicolumn{1}{c}{0.5935}
\\
{} & \multirow{1}{*}{Llama3.2-RAT} & \multirow{1}{*}{3B} & \multicolumn{1}{c}{0.2315} & \multicolumn{1}{c}{0.5996} & \multicolumn{1}{c}{0.6948} & \multicolumn{1}{c}{0.1882} & \multicolumn{1}{c}{0.6061} & \multicolumn{1}{c}{0.6125}
\\ \hline
{} & \multirow{1}{*}{Prefix-Tuning} & \multirow{1}{*}{3B} & \multicolumn{1}{c}{0.3497} & \multicolumn{1}{c}{0.6735} & \multicolumn{1}{c}{0.7062} & \multicolumn{1}{c}{0.2175} & \multicolumn{1}{c}{0.6258} & \multicolumn{1}{c}{0.6182}
\\
{} & \multirow{1}{*}{LoRA-Tuning} & \multirow{1}{*}{3B} & \multicolumn{1}{c}{0.3914} & \multicolumn{1}{c}{0.6920} & \multicolumn{1}{c}{0.7217} & \multicolumn{1}{c}{0.2517} & \multicolumn{1}{c}{0.6319} & \multicolumn{1}{c}{0.6225}
\\
{Tuning-SLM} & \multirow{1}{*}{RAG-LoRA} & \multirow{1}{*}{3B} & \multicolumn{1}{c}{0.3987} & \multicolumn{1}{c}{0.6834} & \multicolumn{1}{c}{0.7240} & \multicolumn{1}{c}{0.2443} & \multicolumn{1}{c}{0.6454} & \multicolumn{1}{c}{0.6342}
\\
{} & \multirow{1}{*}{\textbf{\texttt{EvidenceMap}}} & \multirow{1}{*}{\textbf{1B}} & \multicolumn{1}{c}{\underline{0.4245}} & \multicolumn{1}{c}{\underline{0.7175}} & \multicolumn{1}{c}{0.7311} & \multicolumn{1}{c}{\underline{0.2548}} & \multicolumn{1}{c}{\underline{0.6513}} & \multicolumn{1}{c}{0.6375}
\\
{} & \multirow{1}{*}{\textbf{\texttt{EvidenceMap}}} & \multirow{1}{*}{\textbf{3B}} & \multicolumn{1}{c}{\textbf{0.4429}} & \multicolumn{1}{c}{\textbf{0.7287}} & \multicolumn{1}{c}{\textbf{0.7509}} & \multicolumn{1}{c}{\textbf{0.2931}} & \multicolumn{1}{c}{\textbf{0.6783}} & \multicolumn{1}{c}{\textbf{0.6610}}
\\ \bottomrule
\end{tabular}
\caption{The comparison demonstrates the effectiveness of \textbf{\texttt{EvidenceMap}} across two biomedical datasets. In the inference settings, it surpasses RAG and CoT augmented methods with both the same and larger sizes of models. Our framework also outperforms all the tuning-based methods, even with a smaller generative model.}
\label{performane}
\end{table*}

The results of the performance comparison are shown in Table \ref{performane}. We can make the following observations.

First, \textbf{Learning evidence analysis significantly improves the ability to utilize diverse information.} When using the same generative language model (Llama-3.2-3B), \textbf{\texttt{EvidenceMap}} outperforms RAG, CoT and RAT, demonstrating that our approach effectively leverages the relationships among multiple pieces of evidence and the question, thereby demonstrating the advantages of multifaceted analysis.

Second, \textbf{Learning evidence analysis unleashes the power of small language models to resolve biomedical questions.} Compared to larger models, including the RAG method with general LLMs (GPT-4o-mini, Llama-3.1-8B) and a biomedical LLM (OpenBioLLM-8B), our approach demonstrates superior performance even with a 1B model, indicating that evidence analysis can effectively unleash the ability of smaller language models.

Third, \textbf{Explicitly learning evidence analysis is better than implicit tuning of language models.} We compare three strategies in the setting of model fine-tuning: fine-tuning the small pre-trained model with soft prompt from evidence texts, performance efficient fine-tuning (PEFT) tailored for generative models, and PEFT using prompts derived from multiple pieces of evidence. The results indicate that the performance of these implicit learning strategies is significantly inferior to that of explicit learning of evidence analysis.

\subsection{Ablation Study on Evidence Analysis}
We further investigate the effect of each part in the evidence analysis on overall performance, including supportive evaluation, logical correlation, and content summarization. We systematically remove one part at a time to observe its impact (the percentage of decline denoted as $\Delta$). 

\begin{table}[htbp]
\renewcommand{\arraystretch}{1.2}
\footnotesize
\centering
\begin{tabular}{l|cccc}
\bottomrule
\multirow{2}{*}{\textbf{Method}} &\multicolumn{4}{c}{\textbf{BioASQ}}
\\
& \multicolumn{1}{c}{BERT-S}&\multicolumn{1}{c}{$\Delta$}&\multicolumn{1}{c}{LLM-ACC}&\multicolumn{1}{c}{$\Delta$}
\\ \hline
{w/o Eval} & \multicolumn{1}{c}{0.7085} & \multicolumn{1}{c}{$\downarrow$ 2.78\%} & \multicolumn{1}{c}{0.7326} & \multicolumn{1}{c}{$\downarrow$ 2.44\%}
\\ 
{w/o Cor} & \multicolumn{1}{c}{0.6977} & \multicolumn{1}{c}{$\downarrow$ 4.25\%} & \multicolumn{1}{c}{0.7270} & \multicolumn{1}{c}{$\downarrow$ 3.18\%}
\\ 
{w/o Sum} & \multicolumn{1}{c}{0.7142} & \multicolumn{1}{c}{$\downarrow$ 1.99\%} & \multicolumn{1}{c}{0.7463} & \multicolumn{1}{c}{$\downarrow$ 0.61\%}
\end{tabular}

\begin{tabular}{l|cccc}
\bottomrule
\multirow{2}{*}{\textbf{Method}} &\multicolumn{4}{c}{\textbf{PubMedQA}}
\\
& \multicolumn{1}{c}{BERT-S}&\multicolumn{1}{c}{$\Delta$}&\multicolumn{1}{c}{LLM-ACC}&\multicolumn{1}{c}{$\Delta$}
\\ \hline
{w/o Eval} & \multicolumn{1}{c}{0.6611} & \multicolumn{1}{c}{$\downarrow$ 2.54\%} & \multicolumn{1}{c}{0.6537} & \multicolumn{1}{c}{$\downarrow$ 1.10\%}
\\ 
{w/o Cor} & \multicolumn{1}{c}{0.6556} & \multicolumn{1}{c}{$\downarrow$ 3.34\%} & \multicolumn{1}{c}{0.6518} & \multicolumn{1}{c}{$\downarrow$ 1.39\%}
\\ 
{w/o Sum} & \multicolumn{1}{c}{0.6679} & \multicolumn{1}{c}{$\downarrow$ 1.53\%} & \multicolumn{1}{c}{0.6589} & \multicolumn{1}{c}{$\downarrow$ 0.32\%}
\\ \bottomrule
\end{tabular}
\caption{Ablation study of three aspects within the evidence analysis on two datasets. The results indicate that all aspects of the analysis we proposed have positive impacts on the performance. }
\label{evidence-ablation}
\end{table}

As shown in Table \ref{evidence-ablation}, each part contributes to the overall performance, which is consistent with our hypothesis. In detail, the logical correlation has the most significant impact on overall performance on both datasets, indicating that determining the logical relationship between two pieces of evidence is crucial for providing a high-quality answer. Meanwhile, evaluating how evidence supports the question plays a more significant role than summarizing the evidence, indicating that assessing the relationship between evidence and the question improves effectiveness. Furthermore, incorporating evidence analysis has a greater effect on the BioASQ dataset compared to the PubMedQA dataset in general, showing that the effect of evidence analysis is influenced by the specific data distribution.

\subsection{Ablation Study on Answer Generation}
We further analyze the impact of each module in the answer generation step. We denote the method that does not incorporate textual evidence embeddings as "w/o TE", and the one that eliminates the input of evidence analysis as "w/o EA". We replace the MLP projector with a simple linear layer and denote it as "w/o Proj". Performance results and variations are illustrated in Table \ref{gen-ablation}. The results indicate that both the textual input and our proposed evidence analysis have significant impacts on the final performance, which can dramatically improve the quality of answers beyond the foundation of basic language model generation. In contrast, the projector used to align the evidence analysis results with the generative language model is just an auxiliary to the overall performance.

\begin{table}[htbp]
\renewcommand{\arraystretch}{1.2}
\footnotesize
\centering
\begin{tabular}{l|cccc}
\bottomrule
\multirow{2}{*}{\textbf{Method}} &\multicolumn{4}{c}{\textbf{BioASQ}}
\\
& \multicolumn{1}{c}{BERT-S}&\multicolumn{1}{c}{$\Delta$}&\multicolumn{1}{c}{LLM-ACC}&\multicolumn{1}{c}{$\Delta$}
\\ \hline
{w/o TE} & \multicolumn{1}{c}{0.5977} & \multicolumn{1}{c}{$\downarrow$ 17.98\%} & \multicolumn{1}{c}{0.6156} & \multicolumn{1}{c}{$\downarrow$ 18.02\%}
\\ 
{w/o EA} & \multicolumn{1}{c}{0.6152} & \multicolumn{1}{c}{$\downarrow$ 15.58\%} & \multicolumn{1}{c}{0.6227} & \multicolumn{1}{c}{$\downarrow$ 17.07\%}
\\ 
{w/o Proj} & \multicolumn{1}{c}{0.7115} & \multicolumn{1}{c}{$\downarrow$ 2.36\%} & \multicolumn{1}{c}{0.7442} & \multicolumn{1}{c}{$\downarrow$ 0.89\%}

\end{tabular}
\begin{tabular}{l|cccc}
\bottomrule
\multirow{2}{*}{\textbf{Method}} &\multicolumn{4}{c}{\textbf{PubMedQA}}
\\
& \multicolumn{1}{c}{BERT-S}&\multicolumn{1}{c}{$\Delta$}&\multicolumn{1}{c}{LLM-ACC}&\multicolumn{1}{c}{$\Delta$}
\\ \hline
{w/o TE} & \multicolumn{1}{c}{0.5808} & \multicolumn{1}{c}{$\downarrow$ 14.37\%} & \multicolumn{1}{c}{0.6054} & \multicolumn{1}{c}{$\downarrow$ 8.41\%}
\\ 
{w/o EA} & \multicolumn{1}{c}{0.6099} & \multicolumn{1}{c}{$\downarrow$ 10.08\%} & \multicolumn{1}{c}{0.6384} & \multicolumn{1}{c}{$\downarrow$ 3.42\%}
\\ 
{w/o Proj} & \multicolumn{1}{c}{0.6741} & \multicolumn{1}{c}{$\downarrow$ 0.62\%} & \multicolumn{1}{c}{0.6517} & \multicolumn{1}{c}{$\downarrow$ 1.41\%}
\\ \bottomrule
\end{tabular}
\caption{Ablation study on the answer generation step. Our proposed evidence analysis (EA) demonstrates a significant impact on the effectiveness of answer generation, which is secondary only to that of text input (TE).}
\label{gen-ablation}
\end{table}

\subsection{Impact of Pre-trained Language Models}
One advantage of our framework is that it requires training only a small pre-trained language model to efficiently acquire evidence analysis abilities. We further investigate the impact of various pre-trained language models on overall performance. We selected four representative pre-trained models, with training strategies ranging from masked to casual style, including DistilBERT \cite{sanh2019distilbert} used in our \textbf{\texttt{EvidenceMap}}, and three larger models (BERT-Base \cite{bert-2019}, GPT2 \cite{radford2019language} and ModernBERT \cite{warner2024smarter}). We can observe from the Table \ref{slm-select} that:

\begin{table}[htbp]
\renewcommand{\arraystretch}{1.1}
\footnotesize
\centering
\begin{tabular}{l|cccc}
\bottomrule
\multirow{2}{*}{\textbf{Method}} &\multicolumn{4}{c}{\textbf{BioASQ}}
\\
& \multicolumn{1}{c}{BERT-S}&\multicolumn{1}{c}{$\Delta$}&\multicolumn{1}{c}{LLM-ACC}&\multicolumn{1}{c}{$\Delta$}
\\ \hline
{w/ DistilBERT} & \multirow{2}{*}{0.7287} & \multirow{2}{*}{-} & \multirow{2}{*}{0.7509} & \multirow{2}{*}{-}
\\
{\qquad \textsl{- Base 66M}} & \multicolumn{1}{c}{} & \multicolumn{1}{c}{} & \multicolumn{1}{c}{} & \multicolumn{1}{c}{}
\\ 
{w/ BERT-Base} & \multirow{2}{*}{0.7457} & \multirow{2}{*}{$\uparrow$} & \multirow{2}{*}{0.7441} & \multirow{2}{*}{$\downarrow$}
\\
{\qquad \textsl{- Base 110M}} & \multicolumn{1}{c}{} & \multicolumn{1}{c}{} & \multicolumn{1}{c}{} & \multicolumn{1}{c}{}
\\ 
{w/ GPT2} & \multirow{2}{*}{0.7299} & \multirow{2}{*}{$\uparrow$} & \multirow{2}{*}{0.7576} & \multirow{2}{*}{$\uparrow$}
\\
{\qquad \textsl{- Base 117M}} & \multicolumn{1}{c}{} & \multicolumn{1}{c}{} & \multicolumn{1}{c}{} & \multicolumn{1}{c}{}
\\ 
{w/ ModernBERT} & \multirow{2}{*}{0.7325} & \multirow{2}{*}{$\uparrow$} & \multirow{2}{*}{0.7610} & \multirow{2}{*}{$\uparrow$}
\\
{\qquad \textsl{- Base 149M}} & \multicolumn{1}{c}{} & \multicolumn{1}{c}{} & \multicolumn{1}{c}{} & \multicolumn{1}{c}{}

\end{tabular}
\begin{tabular}{l|cccc}
\bottomrule
\multirow{2}{*}{\textbf{Method}} &\multicolumn{4}{c}{\textbf{PubMedQA}}
\\
& \multicolumn{1}{c}{BERT-S}&\multicolumn{1}{c}{$\Delta$}&\multicolumn{1}{c}{LLM-ACC}&\multicolumn{1}{c}{$\Delta$}
\\ \hline
{w/ DistilBERT} & \multirow{2}{*}{0.6783} & \multirow{2}{*}{-} & \multirow{2}{*}{0.6610} & \multirow{2}{*}{-}
\\
{\qquad \textsl{- Base 66M}} & \multicolumn{1}{c}{} & \multicolumn{1}{c}{} & \multicolumn{1}{c}{} & \multicolumn{1}{c}{}
\\ 
{w/ BERT-Base} & \multirow{2}{*}{0.6872} & \multirow{2}{*}{$\uparrow$} & \multirow{2}{*}{0.6626} & \multirow{2}{*}{$\uparrow$}
\\
{\qquad \textsl{- Base 110M}} & \multicolumn{1}{c}{} & \multicolumn{1}{c}{} & \multicolumn{1}{c}{} & \multicolumn{1}{c}{}
\\ 
{w/ GPT2} & \multirow{2}{*}{0.6814} & \multirow{2}{*}{$\uparrow$} & \multirow{2}{*}{0.6647} & \multirow{2}{*}{$\uparrow$}
\\
{\qquad \textsl{- Base 117M}} & \multicolumn{1}{c}{} & \multicolumn{1}{c}{} & \multicolumn{1}{c}{} & \multicolumn{1}{c}{}
\\ 
{w/ ModernBERT} & \multirow{2}{*}{0.6859} & \multirow{2}{*}{$\uparrow$} & \multirow{2}{*}{0.6746} & \multirow{2}{*}{$\uparrow$}
\\
{\qquad \textsl{- Base 149M}} & \multicolumn{1}{c}{} & \multicolumn{1}{c}{} & \multicolumn{1}{c}{} & \multicolumn{1}{c}{}
\\ \bottomrule
\end{tabular}
\caption{Experiment on the impact of using different small pre-trained language models. DistilBERT, despite having the fewest parameters, still demonstrates competitive capabilities in learning evidence analysis for question answering.}
\label{slm-select}
\end{table}

\textbf{More powerful pre-trained language models can achieve better results.} BERT-Base, GPT2, and ModernBERT, which have larger parameter sizes and better performance in general language abilities, improve results on most indicators. This indicates that as the language model continues to update and advance, the capabilities of our framework will also be further enhanced.

\textbf{Despite having the fewest parameters, DistilBERT remarkably achieves competitive performance.} DistilBERT retrains most of the ability of BERT but is much lighter, which fits well with our framework. With a reduction in parameter size, there is only a slight decline in performance, further illustrating that the evidence analysis we propose is easy to learn in small-scale models.

\subsection{Impact of Evidence Input}
The amount of textual evidence varies in the datasets we used. BioASQ has an average of 26.60 relevant papers, while PubMedQA has an average of only 4.84. To standardize the experimental settings and reduce computational overhead, as mentioned in Section \ref{details}, we set the maximum number of paper acquisitions at 5 and add evidence of LLM to enhance diversity. Meanwhile, as described in Section \ref{acquisition}, we introduce the textual evidence for each question from an LLM to enhance the diversity of input. Therefore, in this relatively complex setting, the impact of the quantity and sources of evidence on performance remains worthy of exploration.

\begin{figure}[htbp]
    \begin{center}
    \includegraphics[width=0.5\textwidth]{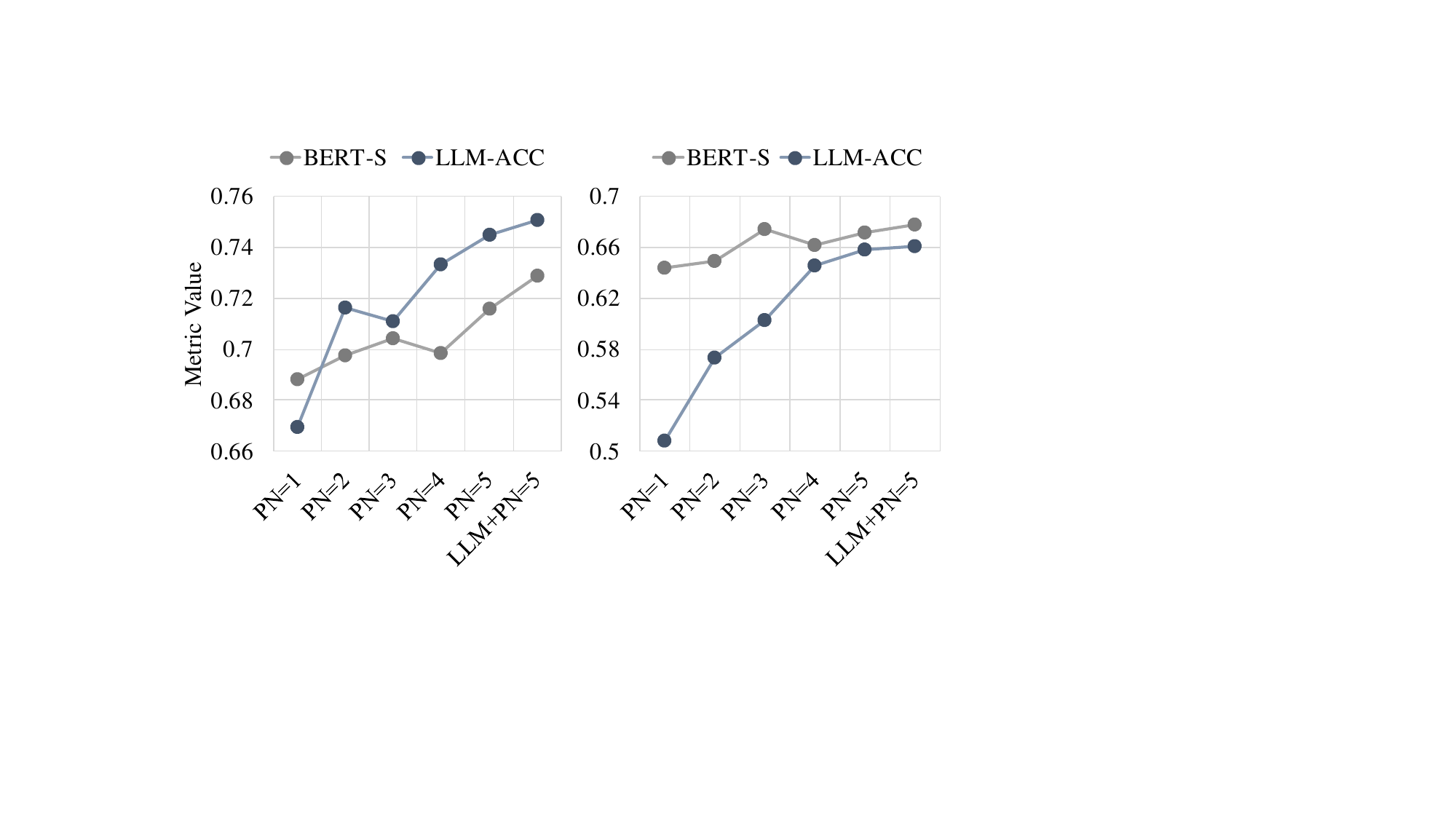} 
    \caption{The variation of BERT-S and LLM-ACC with the number of papers as evidence (PN) and the inclusion of LLM evidence (LLM) on two datasets (Left for BioASQ and right for PubMedQA).}
    \label{evidence-num}
    \end{center}
\end{figure}

Figure \ref{evidence-num} is the chart showing the performance changes with the increase in the number of papers and the addition of evidence from an LLM. In general, overall performance increases with increasing number of papers and LLM-based evidence provided. However, for BioASQ, the rate of increase in the LLM-ACC metric is faster than that of BERTScore and there is a decline only when the number of papers is 3 and 4 for the two metrics, respectively. Although the experiments on PubMedQA demonstrate a lower LLM-ACC, its growth rate is still faster compared to BERTScore. The experimental results on these two datasets indicate that our framework can effectively utilize a greater quantity or a richer source of evidence, thus improving the overall quality of responses.

\section{Case Study}
\begin{figure*}[htbp]
    \begin{center}
    \includegraphics[width=1.0\textwidth]{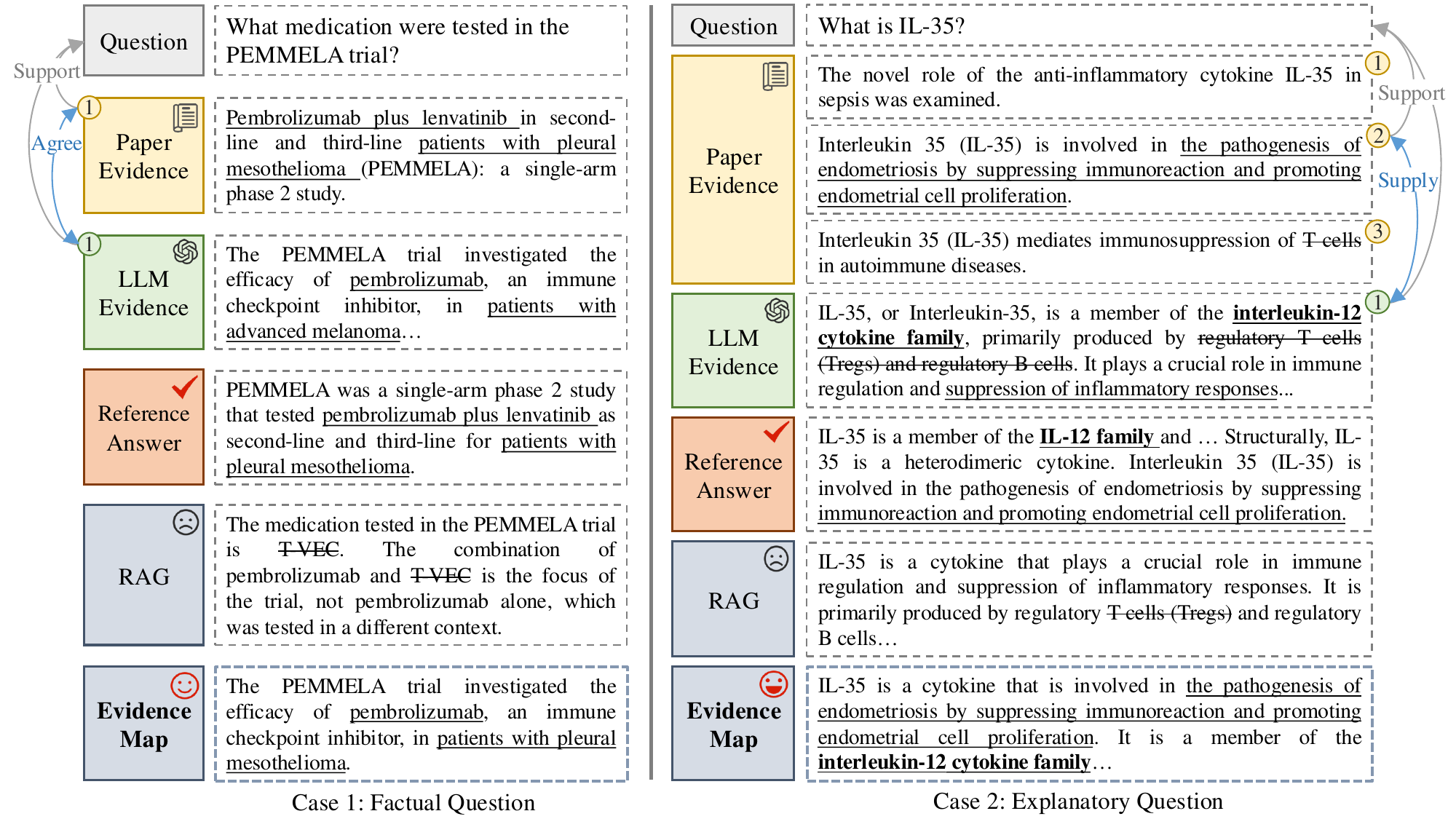} 
    \caption{Two cases with distinct difficulties demonstrate that \textbf{\texttt{EvidenceMap}} can more effectively utilize content from multiple and diverse evidence through learning evidence analysis, thereby providing higher quality answers.}
    \label{case}
    \end{center}
\end{figure*}

In this section, we qualitatively analyze the effectiveness of our framework in addressing biomedical questions by presenting specific cases. Figure \ref{case} presents two examples of utilizing evidence in distinct ways to formulate responses. In each case, a question is provided alongside the relevant evidence, the reference answer, and the answers generated by the 3B version of RAG and our \textbf{\texttt{EvidenceMap}}. 

Case 1 illustrates a scenario of factual questions. We notice that the content of the reference answer is consistent with the evidence presented in the paper, both referencing the drug combination "Pembrolizumab plus lenvatinib", which is more accurate and comprehensive than the LLM evidence. The answer provided by \textbf{\texttt{EvidenceMap}} is incomplete, mentioning only the drug "Pembrolizumab". In contrast, the response generated by the RAG method references "T-VEC", which is an entirely different medication, which results in the dissemination of incorrect information and leads to a decrease in accuracy. This indicates that although our framework may provide incomplete information, analyzing the relationships between pieces of evidence can mitigate the hallucination problem in generative models and help prevent the generation of incorrect answers.

Case 2 illustrates a scenario that addresses an explanatory question. By analyzing the reference answers, we can tell that a more complex analytical process is necessary. Accurately addressing this question requires reliance on information from a piece of paper evidence and the LLM evidence, while simultaneously avoiding certain elements from other papers and the misleading part of the LLM evidence. Specifically, the "IL-12 family" mentioned in the reference answer should be derived from the LLM evidence, which points out the "interleukin-12 cytokine family". Our \textbf{\texttt{EvidenceMap}} accurately captures this information element and presents it in the response. Meanwhile, the reference answer mentions "immunoreaction and promoting endometrial cell proliferation" a concept supported by evidence from the second paper. \textbf{\texttt{EvidenceMap}} precisely identifies this content and incorporates it into the response. In contrast, the answer generated by the RAG method overlooks these key pieces of information and utilizes extraneous content ("T cells") from other papers. This case illustrates that our approach effectively utilizes multiple and diverse sources of evidence by analyzing the relationships between them, thereby providing more accurate answers.

\section{Related Work}
\subsection{Biomedical Evidence Analysis}
Resolving biomedical questions requires a multifaceted analysis of related evidence. A recent study \cite{singhal2025toward} suggests that addressing biomedical questions requires correct retrieval, accurate manipulation, and complete understanding of relevant knowledge. Some studies \cite{jin2021disease,singhal2023large} indicate that models need to ignore irrelevant distractors to avoid incorrect interpretations of the evidence in the answers. Other studies focus on logical reasoning \cite{chen2021logical,kang2023evidencemap} and summarization of evidence \cite{xie2022pre,zhang2024closing}, thereby integrating medical evidence to support decision analysis and provide high-quality answers. Our study builds on these foundational works, integrating the above key aspects of evidence analysis, and proposes a framework for evidence analysis and analysis-augmented generation to address biomedical question-answering tasks.

\subsection{Textual Knowledge Augmented LLM}
Textual evidence, as an external knowledge, can significantly enhance the quality of language model generation in professional domains. Methods for knowledge augmentation can be categorized into two types. The first is the augmentation during inference. RAG \cite{lewis2020retrieval} and Self-RAG \cite{asai2023self} belong to this category, which improve generation effectiveness by retrieving relevant documents as factual knowledge and adjusting these texts during inference. Meanwhile, a model can reduce hallucinations from generation by incorporating reasoning knowledge, as demonstrated by CoT \cite{wei2022chain} and ToT \cite{yao2024tree}, a series of sequential or structured reasoning steps are generated and executed explicitly. Furthermore, methods with an augmentation mixture of factual and reasoning knowledge can further combine the strengths of both, such as RAT \cite{wang2024rat}. Another type of augmentation is by tuning the model with domain text. Soft prompt tuning \cite{prompt-tuning,2023-tailor} uses implicit learnable vectors to replace explicit tokens as prompts to flexibly adapt to multiple downstream tasks. Instruction tuning \cite{2023-self-instruct,lin2023ra} improves models in knowledge-intensive tasks by constructing specific instruction datasets and learning the utilization of input information and contextual understanding. Compared to the aforementioned methods, our study proposes a novel knowledge augmentation process that prompts the inference of generative models by training the explicit evidence analysis abilities with a small pre-trained language model.

\section{Conclusion}
In this work, we present a novel framework for generative biomedical question answering named \textbf{\texttt{EvidenceMap}}, which efficiently addresses the robustness issues of language model generation with multiple pieces of evidence. Inspired by evidence analysis in the biomedical domain, our framework fine-tunes a small pre-trained language model to learn the analysis of multifaceted aspects of evidence and prompt a small generative model to produce answers. The experimental results and representative cases on public datasets show that \textbf{\texttt{EvidenceMap}} can significantly surpass popular reasoning methods with equivalent and larger language models by fine-tuning a pre-trained model with only 66M parameters.

\section*{Limitations}
We faithfully discuss the limitations that we would like to improve in future work. 

First, we currently evaluate our method with popular long-form metrics on public biomedical datasets. Future studies continue to propose specific and rational metrics for the biomedical domain, while also conducting tests in additional professional domains.

Second, the evidence map and the learning of evidence analysis that we propose provide a guide to improve the ability to use evidence. We will continue to explore enabling the model to analyze fine-grained and symbolic logical relationships among evidence to further improve model robustness while using multiple pieces of evidence.

Third, despite our method demonstrating superior performance in generation with small language models, we will continue to conduct in-depth research on the interpretability and stability of model inference, exploring further advantages in evidence analysis, and with larger language models.

Fourth, the generalizability of learning evidence analysis and the potential transferability to various pre-trained models still remains an area for further exploration, which will justify whether evidence analysis is a universal competency, and further improve the efficiency of applying evidence analysis.

\section*{Ethical Considerations}
Our \textbf{\texttt{EvidenceMap}} framework is not flawless due to limited model capabilities and the possibility of errors or outdated evidence. We test it on two public datasets, BioASQ and PubMedQA, and acknowledge that its effectiveness may be restricted to similar datasets or domains. Its performance in other scenarios is uncertain and poses potential risks. Therefore, it is crucial to exercise caution and verify the accuracy of the answers generated by the method.


\bibliography{custom}

\appendix

\section{Prompts Used in EvidenceMap}\label{appendix-prompt}

\subsection{Prompts for Evidence Map} \label{prompt-evidence}
The $\texttt{prompt}_i$ to obtain the representation of an individual evidence $X_p^i$ from an small language model:

\begin{center}
\fcolorbox{black}{gray!10}{
    \parbox{1.0\linewidth}{Evidence: $E_p^i$ \\ This evidence means: }
}
\end{center}

The $\texttt{prompt}_e$ to obtain the representation of the supportive evaluation of evidence $R_{eval}^i$ from an small language model:

\begin{center}
\fcolorbox{black}{gray!10}{
    \parbox{1.0\linewidth}{Evidence: $E_p^i$ \\ Question: $Q$ \\ How much the evidence supports the question:}
}
\end{center}

The $\texttt{prompt}_c$ to obtain the representation of the logical correlation of two pieces of evidence $R_{cor}^{(i,j)}$ from an small language model:

\begin{center}
\fcolorbox{black}{gray!10}{
    \parbox{1.0\linewidth}{Evidence1: $E_p^i$ \\ Evidence2: $E_p^j$ \\ The logical relationship between these two pieces of evidence:}
}
\end{center}

\subsection{Prompt for Acquiring LLM Evidence} \label{prompt-acquire}
To simulate the sources of evidence in real problem-solving scenarios, we introduced evidence texts based on the intrinsic knowledge of LLMs, thereby enriching the evidence content. This evidence is analyzed alongside the paper evidence to explore its impact on performance. We prompt GPT-4o to utilize its intrinsic knowledge to acquire evidence for each question as follows.

\begin{center}
\fcolorbox{black}{gray!10}{
    \parbox{1.0\linewidth}{
    \textit{System Input:} You are an AI assistant that helps a human analyst discover evidence that supports the question.
    \vskip .2cm
    \textit{User Input:} Provide a concise summary that supports answering the question with your own knowledge. The summary should be evidence including key insights that can explain your answer to the question. \\
    Question: $Q$ \\
    Summary:}
}
\end{center}

\subsection{Prompt for LLM-based Metric} \label{prompt-llm}
We prompt GPT-4o to provide both the accuracy score and the fluency score ranging from 0.0 to 1.0 as follows.

\begin{center}
\fcolorbox{black}{gray!10}{
    \parbox{1.0\linewidth}{
    \textit{System Input:} You are an AI assistant tasked with evaluating the quality of the generated answers in terms of accuracy and fluency. 
    \vskip .2cm
    \textit{User Input:} Please score the generated answer based on accuracy and fluency separately, comparing with the reference answer. \\
    The scores should be a float number between 0.0 and 1.0, where 0.0 indicates the generated answer is completely incorrect or unable to understand, 1.0 means the generated answer is totally precise and fluent.\\
    Your output should be a dictionary such as \{"accuracy": 0.8, "fluency": 0.9\}, without any additional information.\\
    The question, generated answer, and reference answer are give as below:\\
    Question: $Q$\\
    Generated answer: $A$\\
    Reference answer: $A^*$\\
    Your output:}
}
\end{center}

\section{Details of Baseline Settings} \label{baseline}
We implement baseline methods by applying various knowledge augmentation strategies based on different pre-trained models for comparison with our framework. For the \textit{Inference-LLM} setting, we provide evidence the same as EvidenceMap for each question to perform inference on GPT-4o-mini and Llama3.1-8B, implementing a naive RAG approach. We also directly evaluate the adapted OpenBioLLM-8B, which have been tuned by domain datasets. For the \textit{Inference-SLM} setting, we implement a CoT process based on Llama3.2-3B, providing the three phases of evidence analysis in natural language for the model to follow. In addition, a RAT framework is implemented, enabling the model to generate an initial CoT and optimize the thinking process based on acquired evidence and ultimately perform inference based on the adjusted reasoning process. For the \textit{Tuning-SLM} setting, we perform prefix-tuning by providing the Llama3.2 model with a soft prompt, which can be optimized with the pre-trained BERT-like model. We also achieve instruction-tuning strategy by constructing evidence-augmented instructions as the dataset, and update Llama3.2-3B using LoRA, which is denoted as RAG-LoRA.

\section{Details of Biomedical Datasets}
The number of samples and the average amount of evidence per sample in the training and test sets from the two public biomedical datasets are shown in Table \ref{dataset}. The obvious difference in the number of evidence and the results shown in Table \ref{performane} indicates that our method can be fitted with variable data conditions.

\begin{table}[htbp]
\renewcommand{\arraystretch}{1.2}
\footnotesize
\centering
\begin{tabular}{l|ccc}
\bottomrule
\multirow{1}{*}{\textbf{Dataset}} & \multicolumn{1}{c}{\textbf{Set}} & \multicolumn{1}{c}{\textbf{Samples}} & \multicolumn{1}{c}{\textbf{Evidence per Sample}}
\\ \hline
\multirow{2}{*}{BioASQ}  & \multicolumn{1}{c}{Train} & \multicolumn{1}{c}{5049} & \multicolumn{1}{c}{12.91}
\\ 
 & \multicolumn{1}{c}{Test} & \multicolumn{1}{c}{340} & \multicolumn{1}{c}{26.60}
\\ 
\multirow{2}{*}{PubMedQA}  & \multicolumn{1}{c}{Train} & \multicolumn{1}{c}{953} & \multicolumn{1}{c}{4.35}
\\ 
 & \multicolumn{1}{c}{Test} & \multicolumn{1}{c}{47} & \multicolumn{1}{c}{4.84}
\\ \bottomrule
\end{tabular}
\caption{Statistical data from the two datasets reveal obvious disparities in the number of samples and average amount of evidence.}
\label{dataset}
\end{table}

\section{Results on More Generative Models}
\begin{figure}[htbp]
    \begin{center}
    \includegraphics[width=0.46\textwidth]{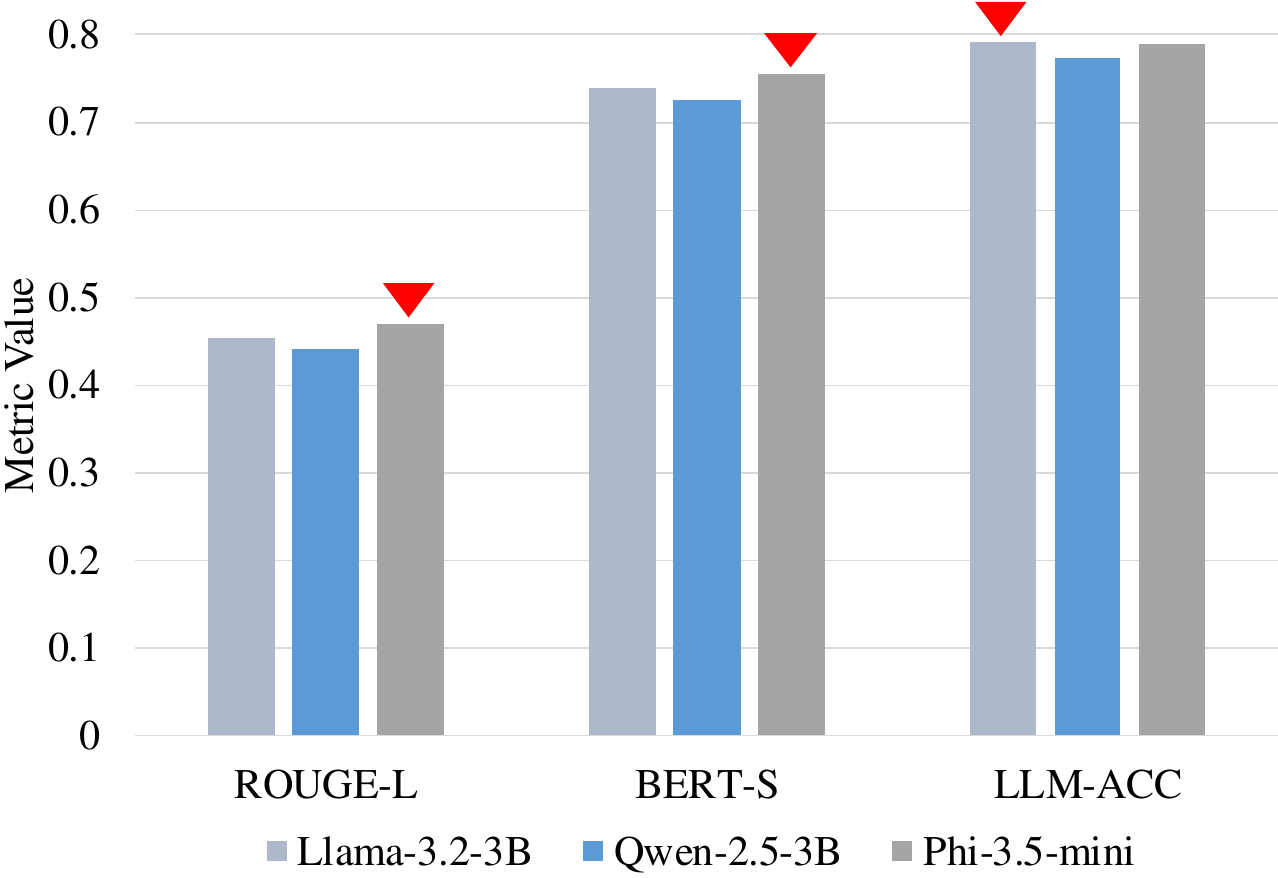} 
    \caption{Results on BioASQ with various small generative language models which have parameter size around 3B. The best scores of three metrics are highlighted.}
    \label{gen-lm}
    \end{center}
\end{figure}

We also evaluate our method on more small generative models around the 3B parameter size, including Qwen-2.5-3B-Instruct \cite{yang2024qwen2} and Phi-3.5-Mini-Instruct \cite{abdin2024phi}. The results of experiments on BioASQ dataset shown in Figure \ref{gen-lm} demonstrate that the performance of \textbf{\texttt{EvidenceMap}} can remain consistent across a wider range of small generative models.

\section{Results of Language Fluency}
We explore the fluency of the language generated by our framework to verify whether the method preserves the original language generation capabilities. As stated in Appendix \ref{prompt-llm}, an LLM is employed to score accuracy and fluency of the generated output at the same time. The average values of all samples from the test set are calculated and are shown in Table \ref{fluency}. The results indicate that our framework achieves the highest and competitive fluency on BioASQ and PubMedQA, respectively, while improving the quality of the answers.
\begin{table}[htbp]
\renewcommand{\arraystretch}{1.2}
\footnotesize
\centering
\begin{tabular}{lccc}
\bottomrule
\multirow{2}{*}{\textbf{Method}}&\multirow{2}{*}{\textbf{Size}}&\multicolumn{1}{c}{\textbf{BioASQ}} & \multicolumn{1}{c}{\textbf{PubMedQA}}
\\ 
& \multicolumn{1}{l}{} & \multicolumn{1}{c}{LLM-FLU} & \multicolumn{1}{c}{LLM-FLU}
\\ \hline   
\multirow{1}{*}{4o-mini-RAG} & \multirow{1}{*}{-} & \multicolumn{1}{c}{0.8795} & \multicolumn{1}{c}{0.8323}
\\
\multirow{1}{*}{Llama3.1-RAG} & \multirow{1}{*}{8B} & \multicolumn{1}{c}{0.8738} & \multicolumn{1}{c}{0.8574}
\\
\multirow{1}{*}{OpenBioLLM} & \multirow{1}{*}{8B} & \multicolumn{1}{c}{0.8662} & \multicolumn{1}{c}{0.8643}
\\ \hline 
\multirow{1}{*}{Llama3.2-RAG} & \multirow{1}{*}{3B} & \multicolumn{1}{c}{\underline{0.8910}} & \multicolumn{1}{c}{\textbf{0.8819}}
\\
\multirow{1}{*}{Analysis-CoT} & \multirow{1}{*}{3B} & \multicolumn{1}{c}{0.8547} & \multicolumn{1}{c}{0.8541}
\\
\multirow{1}{*}{Llama3.2-RAT} & \multirow{1}{*}{3B} & \multicolumn{1}{c}{0.8719} & \multicolumn{1}{c}{0.8675}
\\ \hline
\multirow{1}{*}{Prefix-Tuning} & \multirow{1}{*}{3B} & \multicolumn{1}{c}{0.8592} & \multicolumn{1}{c}{0.8245}
\\
\multirow{1}{*}{LoRA-Tuning} & \multirow{1}{*}{3B} & \multicolumn{1}{c}{0.8558} & \multicolumn{1}{c}{0.8374}
\\
\multirow{1}{*}{RAG-LoRA} & \multirow{1}{*}{3B} & \multicolumn{1}{c}{0.8734} & \multicolumn{1}{c}{0.8422}
\\
\multirow{1}{*}{\textbf{\texttt{EvidenceMap}}} & \multirow{1}{*}{\textbf{1B}} & \multicolumn{1}{c}{0.8829} & \multicolumn{1}{c}{0.8451}
\\
\multirow{1}{*}{\textbf{\texttt{EvidenceMap}}} & \multirow{1}{*}{\textbf{3B}} & \multicolumn{1}{c}{\textbf{0.9088}} & \multicolumn{1}{c}{\underline{0.8746}}
\\ \bottomrule
\end{tabular}
\caption{The comparison of language fluency measured by LLM on two datasets. Our language model training architecture saves generation quality on fluency comparing with other language model based methods.}
\label{fluency}
\end{table}

\section{Complexity Analysis}
The complexity of the framework arises primarily from the combinatorial analysis of multiple evidence during training. Specifically, if there are $N$ pieces of evidence, it is necessary to evaluate the supportive relationship of each piece of evidence to the question, which involves representing $N$ relationships. Furthermore, it is essential to determine the logical relationships between two pieces of evidence, resulting in the representation of $\frac{N\times(N-1)}{2}$ relationships. In practice, we set the number of evidence as 5 and the batch size as 4, which allows the entire training process to run on a single GPU. In addition, conducting evidence quality assessment sequentially or selecting more efficient parallel encoders can further reduce complexity.

\end{document}